\documentclass[10pt,twocolumn,letterpaper]{article}

\usepackage{cvpr}
\usepackage{times}
\usepackage{epsfig}
\usepackage{graphicx}
\usepackage{amsmath}
\usepackage{amssymb}
\usepackage[dvipsnames]{xcolor}
\usepackage{arydshln}
\usepackage{booktabs}
\usepackage{footmisc}

\usepackage[breaklinks=true,bookmarks=false]{hyperref}

\cvprfinalcopy 

\def\cvprPaperID{1115} 

\newcommand{\todo}[1]{}
\renewcommand{\todo}[1]{{\color{red} TODO: {#1}}}

\ifcvprfinal\fi
\begin{document}

\title{Query-guided End-to-End Person Search}

\author{Bharti Munjal\textsuperscript{1,2}
\qquad
Sikandar Amin\textsuperscript{1}
\qquad
Federico Tombari\textsuperscript{2}
\qquad
Fabio Galasso\textsuperscript{1} \\
\textsuperscript{1}OSRAM GmbH, \textsuperscript{2}Technische Universit{\"a}t M{\"u}nchen
}

\maketitle

\thispagestyle{empty}

\begin{abstract}
Person search has recently gained attention as the novel task of finding a person, provided as a cropped sample, from a gallery of non-cropped images, whereby several other people are also visible.
We believe that \textbf{i.}\ person detection and re-identification should be pursued in a joint optimization framework and that \textbf{ii.}\ the person search should leverage the query image extensively (e.g.\ emphasizing unique query patterns). However, so far, no prior art realizes this.

We introduce a novel \emph{query-guided end-to-end person search} network (QEEPS) to address both aspects. We leverage a most recent joint detector and re-identification work, OIM~\cite{xiao2017joint}. We extend this with \textbf{i.}\ a query-guided Siamese squeeze-and-excitation network (QSSE-Net) that uses global context from both the query and gallery images, \textbf{ii.}\ a query-guided region proposal network (QRPN) to produce query-relevant proposals, and \textbf{iii.}\ a query-guided similarity subnetwork (QSimNet), to learn a query-guided re-identification score.
QEEPS is the first end-to-end query-guided detection and re-id network.
On both the most recent CUHK-SYSU~\cite{xiao2017joint} and PRW~\cite{zheng2016prw} datasets, we outperform the previous state-of-the-art by a large margin.


\end{abstract}

\section{Introduction}

Person search has recently emerged as the task of finding a person, provided as a cropped exemplar, in a gallery of non-cropped images~\cite{Liu2017NPSM,xiao2017joint,Xu2014PSS,zheng2016prw}. Person search is challenging, since the gallery contains cluttered background (including additional people) and occlusion. Furthermore, the query person may appear in the gallery under different viewpoints, poses, scale and illumination conditions. However the task is of great relevance in video surveillance, since it enables cross-camera visual tracking~\cite{Castaeda2016ScalableSA} and person verification~\cite{Yang2016LargeSS}.

\begin{figure}[t] 
\begin{center}
	\includegraphics[trim=0cm 0cm 0cm 0cm, clip=true, width=1.0\linewidth]{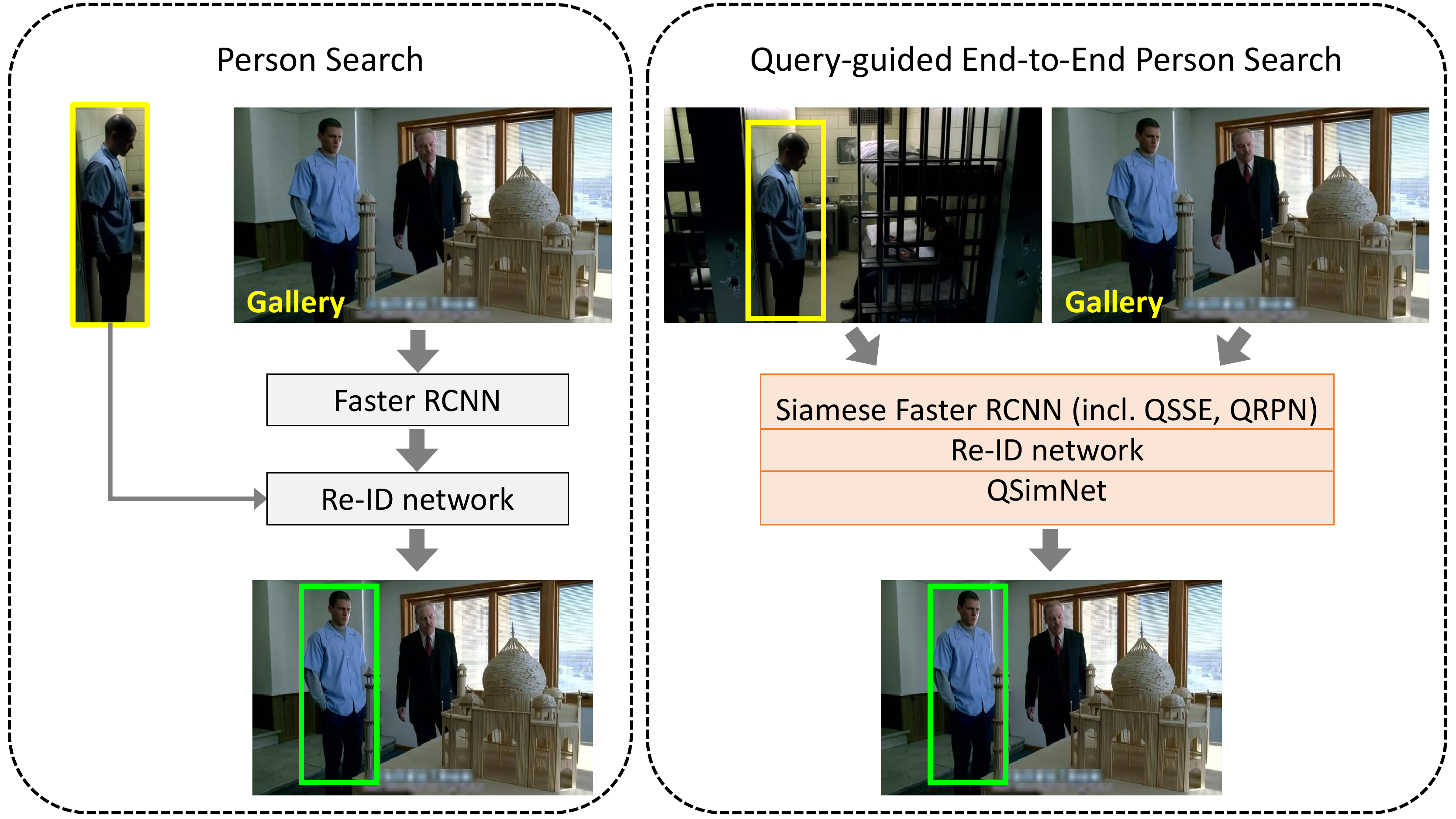}
	
	\caption{Person search is about finding a query person (yellow rectangle) within a gallery image (the target green rectangle). (\textit{Left}) current approaches detect all people from the gallery image, then estimate re-identification features to match the cropped query. (\textit{Right}) our proposed QEEPS guides the person search with an extensive use of the full query image, by means of a joint detection and re-identification network, which is end-to-end trained.
	}
	\vspace{-0.5cm}
	\label{fig:teaser}
\end{center}
\end{figure}

Typical approaches to person search separate the problem into person localization (detection) and re-identification (re-id), and tackle each task sequentially via separate supervised networks. One such approach is the current best performer Mask-G~\cite{Chen_2018_ECCV}. But when separating the two tasks, one may remove useful contextual information for the re-id network (e.g.\ in Fig.\ref{fig:teaser}\textit{(left)} the query is cut out from the query image).
Also, the detection and localization task cannot exploit the information from the query, since the detection network runs independently, before the re-id network.

If we, as humans, were to search for a person in an image, we would not only look at each individual, but we would also search for peculiar patterns in the image, \eg a distinct color or the texture of the person's t-shirt, as an additional hints.
Motivated by this perspective, we introduce the first \emph{Query-guided End-to-End Person Search} work (QEEPS).
We propose the joint optimization of detection and re-identification, and to condition both aspects on the given query image, as exemplified in Fig.\ref{fig:teaser}\textit{(right)}.


Our approach is the only method which is both end-to-end and query-guided. To the best of our knowledge, across the person search approaches~\cite{Liu2017NPSM,xiao2017joint,Xu2014PSS,Zhang2015BeyondFF,zheng2016prw}, only OIM~\cite{xiao2017joint} and IAN~\cite{Xiao2017IANTI} optimize jointly the detector and the re-identification networks (end-to-end).
On the other hand, NPSM~\cite{Liu2017NPSM} is the sole to adopt a query attention mechanism, by replacing the detector RPN with an iterative query-guided search based on Conv-LSTM. In fact NPSM builds on OIM but it is not end-to-end, since it freezes the detector and re-identification network parts to pre-trained values, \ie its re-identification score (used for matching) does not change from the original OIM value.

We are inspired by OIM to design a model encompassing a detector with an additional re-identification branch. As in OIM, we optimize the networks jointly by adopting an OIM loss function~\cite{xiao2017joint}.
Our model additionally features a Query-guided Siamese Squeeze-and-Excitation Network (QSSE-Net), a Query-guided RPN (QRPN) and a Query-Similarity Network (QSimNet).
The QSSE-Net extends the recent squeeze-and-excitation (SE) technique to re-calibrate the channel-wise Siamese feature responses by the (global, image-level) inter-dependencies of the query and gallery channels~\cite{Hu_2018_CVPR}.
The QRPN complements the parallel RPN with query-specific proposals. It employs a modified SE block to emphasize spatial features (in addition to feature channels), particularly bringing up the query-specific discriminant ones.
QSimNet takes the query and gallery image proposal re-id features and provides a query-guided re-id score. When added to the baseline, QSimNet alone improves the mAP and CMC top-1 (hereinafter referred to as top-1) performances by as much as 7.1pp and 4.3pp on the CUHK-SYSU dataset~\cite{xiao2017joint} (gallery size of 100).

Altogether, QEEPS sets a novel state-of-the-art performance of 88.9\% mAP and 89.1\% top-1 on the CUHK-SYSU dataset~\cite{xiao2017joint}, outperforming the prior best performer Mask-G~\cite{Chen_2018_ECCV} by 5.9pp mAP and 5.4pp top-1. Similarly, on the PRW dataset~\cite{zheng2016prw}, QEEPS outperforms Mask-G~\cite{Chen_2018_ECCV} by 4.5pp mAP and 4.6pp top-1 setting state-of-the-art performance of 37.1\% mAP and 76.7\% top-1.

We summarize our contributions: \textbf{i.}\ we introduce the first query-guided end-to-end person search (QEEPS) network; \textbf{ii.}\ we propose a query-guided Siamese squeeze-and-excitation (QSSE) block that extends the interaction between feature channels to additionally model the global similarities between the query and gallery image pairs; \textbf{iii.}\ we define a novel query-guided RPN (QRPN), by extending the SE-Net squeeze-and-excitation block to the query channels and spatial features; \textbf{iv.}\ we define a novel query-similarity subnetwork (QSimNet) to learn a query-guided re-identification score; \textbf{v.}\ we achieve a new state-of-the-art performance on CUHK-SYSU~\cite{xiao2017joint} and PRW~\cite{zheng2016prw} datasets.

\section{Related Work}\label{sec:related}

In this section we first review prior art on the two separate tasks of person detection and person re-identification. Then we review literature on person search

%

\textbf{Person Detection.}
In the past few decades, this field has witnessed steep improvement with the introduction of boosting~\cite{Viola2001RapidOD}, deformable parts models~\cite{Felzenszwalb2009ObjectDW} and aggregate channel features~\cite{Dollr2014FastFP}. 
Convolutional neural networks (CNNs) excel today at this task thanks to jointly learning the classification model and the features~\cite{ren2015faster}, in an end-to-end fashion.
While single-stage object detectors~\cite{Lin2017FocalLF,Liu2016SSDSS,Redmon2017YOLO9000BF} are preferable for runtime performance, the two-stage strategy of Faster R-CNN remains the more robust general solution~\cite{kaiming2017mask}, versatile to tailor region proposals to custom scene geometries~\cite{Amin2017GeometricPF} and to add multi-task branches~\cite{hasan2017tiny,xiao2017joint}. As in OIM~\cite{xiao2017joint}, we adopt Faster R-CNN with a ResNet~\cite{he2016resnet} backbone. 

\textbf{Person Re-Identification.}
%
Classic approaches for person re-identification have focused on manual feature design~\cite{Wang2007ShapeAA,Gray2008ViewpointIP,Farenzena2010PersonRB,Zhao2013UnsupervisedSL} and metric learning~\cite{Liao2015PersonRB,Zhao2017PersonRB,Kstinger2012LargeSM,Li2015MultiScaleLF,Liao2015EfficientPC,Paisitkriangkrai2015LearningTR,Ali_2018_ECCV}.
As in object detection, CNNs have recently conquered the scene in re-identification, too~\cite{ahmed2015cvpr,li2014cvpr_deepreid,Yi2014DeepML}.

While modern CNN approaches target the estimation of a re-id embedding space (whereby the same IDs lie close and further from other individuals), there are two main trends in the model learning: \textbf{i.}\ by Siamese networks and contrastive losses; and \textbf{ii.}\ by ID classification with cross-entropy losses.
In the first, pairs~\cite{ahmed2015cvpr,li2014cvpr_deepreid,Liu2017EndtoEndCA,Varior2016ASL,Xu2018AttCompNet,Yi2014DeepML}, triplets~\cite{Cheng2016PersonRB,Ding2015DeepFL} or quadruplets~\cite{Chen2017BeyondTL} are used to learn a corresponding number of Siamese networks, by pushing or pulling the same or the different person ids, respectively.
In the second, \cite{Xiao2016LearningDF,Zheng2016MARSAV} define as many classes as people IDs, train classifiers with a cross-entropy loss, and take the network features as the embedding metric during inference. Best performing person search approaches~\cite{Liu2017NPSM,xiao2017joint} follow this second trend, which we also adopt.

Our work also relates to the similarity-guided graph neural network of \cite{Shen_2018_ECCV}. They learn the similarity among multiple query and gallery identities and use it to construct a graph, as opposed to a fixed metric, such as the cosine similarity in OIM~\cite{xiao2017joint}. Here we learn the similarity but do not adopt graphs, thus preserving a convenient runtime.

\begin{figure*}[t!] 
\begin{center}
	\includegraphics[trim=0cm 7.5cm 0cm 0cm, clip=true, width=1.0\linewidth]{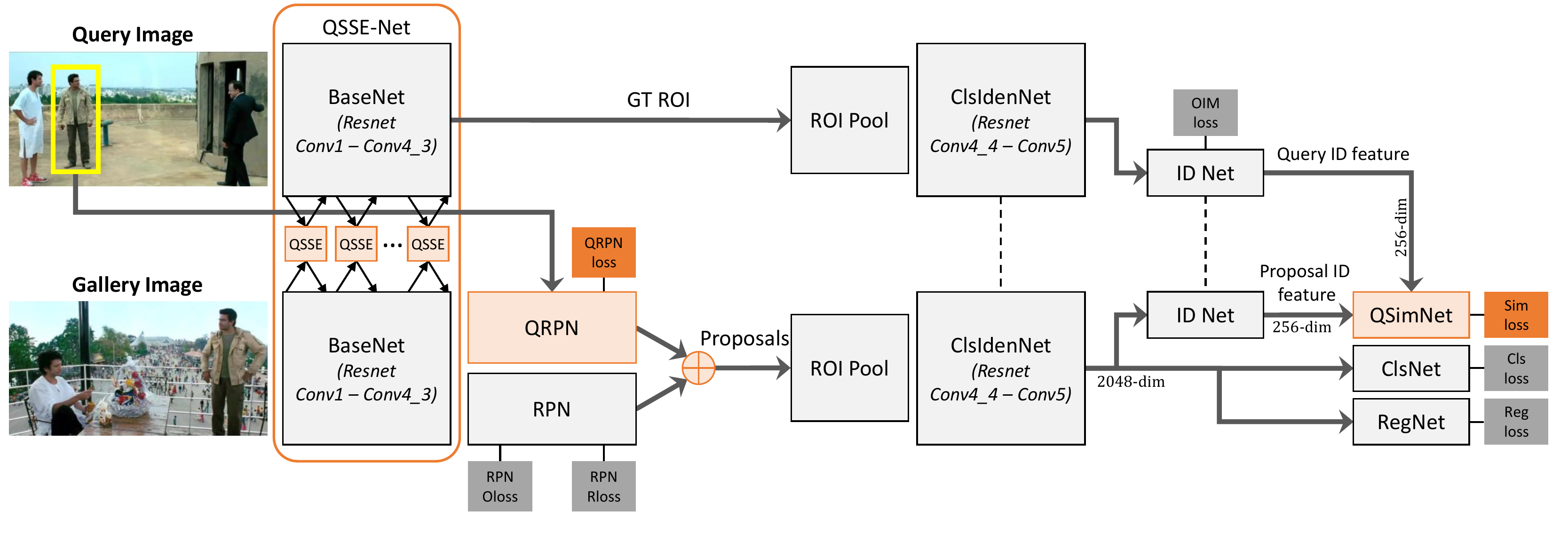}
    \vspace{-0.4cm}
	\caption{
	Our proposed QEEPS network architecture. We pair the reference OIM~\cite{xiao2017joint} \textit{bottom network} with a novel Siamese \textit{top network}, to process the query and guide the person search at different levels of supervision (cf.\ Sec.~\ref{sec:methodoverview}).
	The novel query-guidance blocks of our approach, displayed in orange, are trained end-to-end with the whole network with specific loss functions (\textit{darker orange boxes}).
	}\label{fig:network}
	\vspace{-0.5cm}
\end{center}
\end{figure*}

%
%

\textbf{Person Search.}
The pioneering work of Xu~\etal~\cite{Xu2014PSS} introduces person search as re-identifying people within gallery images, where they also have to be detected and localized. The adoption of CNNs in person search is enabled by the introduction of two recent person search datasets, PRW~\cite{zheng2016prw} and CUHK-SYSU~\cite{xiao2017joint}. Initial approaches~\cite{Xu2014PSS,zheng2016prw} use separate pre-trained people detectors and only learn re-identification networks. Interestingly, the most recent work to date~\cite{Chen_2018_ECCV} states that detection and re-identification should be addressed separately for best performance. We contrast this statement by showing that our single end-to-end network yields better performance.


\textbf{End-to-End Person Search.}
Xiao~\etal~\cite{xiao2017joint} introduces the first end-to-end person detection and re-identification network. 
They propose an Online Instance Matching (OIM) loss to address the challenge of training a classifier matrix for an overwhelming number of person IDs (thousands of different people), as required for both the CUHK-SYSU~\cite{xiao2017joint} and PRW~\cite{zheng2016prw} person search datasets. In other words, they build-up a matrix look-up table by leveraging the IDs in each mini-batch at training, instead of learning ID-specific classifiers. The look-up tables are updated during training by running averages and allow for employing a soft-max loss in the ID Net training with a limited number of IDs.
More recently, \cite{Xiao2017IANTI} extends the OIM with an additional center loss~\cite{Wen2016ADF}, which improves the intra-class feature compactness.
To our knowledge, the OIM loss is currently best for optimizing the joint network, adopted by most recent work~\cite{Liu2017NPSM,Xiao2017IANTI}, including ours.



\textbf{Query-guided person search.}
To the best of our knowledge, the NPSM approach of Liu~\etal~\cite{Liu2017NPSM} is the sole to exploit the query image. They do so by instantiating an iterative person search mechanism based on a Conv-LSTM, which re-weights attention across a number of pre-defined person detection proposals. NPSM builds upon Faster-R-CNN and OIM, but replaces the traditional RPN with the attention mechanism. We note that both the base Faster-R-CNN network and the re-identification head are pre-trained as from \cite{xiao2017joint} and frozen. This implies that, upon the query-guided attention search, the final re-id score remains the same as in \cite{xiao2017joint}, not profiting from the proposal adjustment. We adopt the same Faster-R-CNN network and OIM loss, but optimize those end-to-end, alongside our novel query-guided proposal network.

\section{Background - Online Instance Matching}
We leverage the end-to-end person search architecture of \cite{xiao2017joint}, which we refer to as \textbf{OIM} hereinafter, since it introduces the Online Instance Matching, key to the joint detection and re-identification optimization (cf.\ Sec.~\ref{sec:related}).

We illustrate the base architecture of \cite{xiao2017joint} in Fig.~\ref{fig:network} (\textit{gray blocks} from the \textit{bottom network}, applied to the gallery image).
The OIM network consists of a Faster R-CNN~\cite{ren2015faster} with a ResNet backbone~\cite{he2016resnet} (this accounts for the blocks BaseNet, RPN, ROI Pool and ClsIdenNet in Fig.~\ref{fig:network}). In parallel to the classification (ClsNet) and regression (RegNet) branches, \cite{xiao2017joint} defines an ID Net, which provides a re-identification feature embedding, supposedly unique for the same identities but different for other people. Then they adopt cosine similarity to match cropped query identities to the estimated id embeddings from the gallery image.


\section{Query-guided Person Search}\label{sec:methodoverview}

Fig.~\ref{fig:network} illustrates our proposed architecture.
In more detail, we pair the OIM network~\cite{xiao2017joint}, originally employed for the gallery image, with a second Siamese network (\textit{top network} in Fig.~\ref{fig:network}), applied to the query image. The query network shares weights with the gallery image network. Features from the Siamese query network are used to guide the gallery image network (\textit{bottom network} in Fig.~\ref{fig:network}) at different levels of supervision (novel query-guidance blocks are represented in \textit{orange}).

In more details, we introduce 3 novel subnetworks: \textbf{i.}\ a Query-guided Siamese Squeeze-and-Excitation Network (QSSE-Net) that leverages \emph{global} contextual information from both the query and gallery images to re-weight the feature channels; \textbf{ii.}\ a Query-guided Region Proposal Network (QRPN), leveraging query-ROI-Pooled features to emphasize discriminant patterns in the gallery image to produce relevant proposals; and \textbf{iii.}\ a Query-guided Similarity Network (QSimNet) for computing the re-identification (re-id) jointly from the query and gallery image crop features.

Note that QSSE-Net processes the full query and gallery images and considers therefore a \emph{global} context (e.g.\ if one of the two images is very dark, channels expressing shape are likely to be more discriminant than those encoding color). On the other hand, QRPN and QSimNet are \emph{local}, since they consider the person crop, and dedicated to emphasize features specific to each individual, as defined by the pair [query-gallery] image crop.

\begin{figure}[t] 
\begin{center}
	\includegraphics[trim=0cm 0cm 1cm 0cm, clip=true, width=1.0\linewidth]{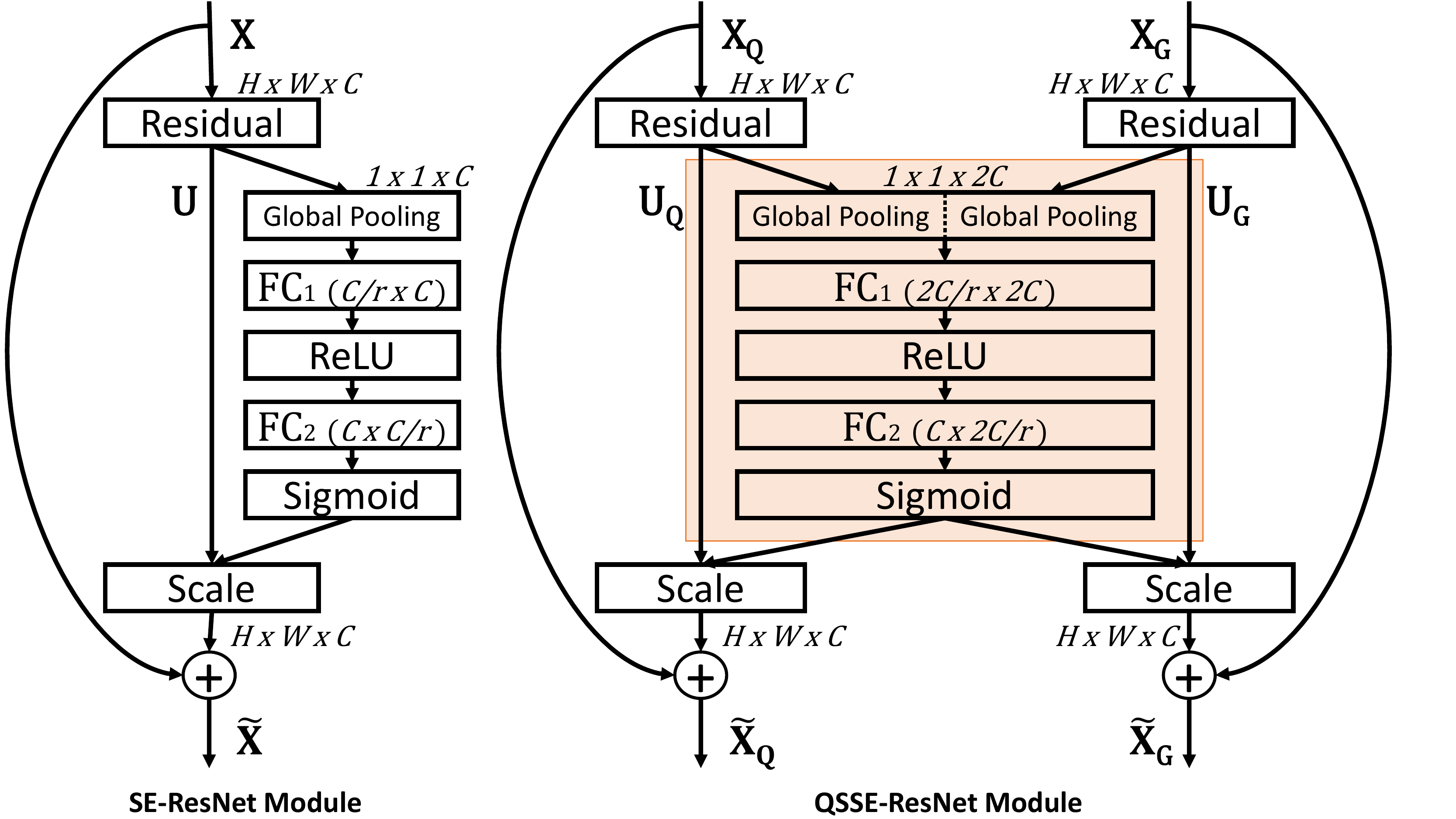}
	\caption{
	Our proposed Query-guided Siamese Squeeze-and-Excitation Network (QSSE-Net).
	QSSE-Net is integrated into the ResNet base network. It concatenates the query and gallery features, upon the residual blocks. It applies then squeeze-and-excitation~\cite{Hu_2018_CVPR}, and re-calibrates the query and the gallery image channels according to intra- and inter-channel dependencies.
	}\label{fig:qseblock}
    \vspace{-0.5cm}
	
\end{center}
\end{figure}

\subsection{Query-guided Siamese Squeeze-and-Excitation Network (QSSE-Net)}\label{sec:qssenet}

The QSSE-Net block is integrated into the ResNet base network. QSSE-Net is inspired by the recent squeeze-and-excitation network (SE-Net)~\cite{Hu_2018_CVPR}, the main difference being the extension to a Siamese-like model which includes both the query and the gallery, as illustrated in Fig.~\ref{fig:qseblock}. The very recent Mask-G \cite{Chen_2018_ECCV} also utilizes squeeze-and-excitation block in their pipeline to re-weight the feature channels.

As proposed in \cite{Hu_2018_CVPR}, a QSSE block performs two operations, namely \emph{squeeze} and \emph{excitation}, i.e.\ compute a weight vector and re-weight the feature maps per channel.
The squeeze operation condenses the spatial information of each of the $C$ channels of both query and gallery by global average pooling, resulting in channel-descriptors $\mathrm{\bf z}_q$ and $\mathrm{\bf z}_g \in\mathbb{R}^{C}$, respectively.

The excitation operation applies a non-linear bottleneck function of two fully-connected layers using concatenated query and gallery channel-descriptors $[\mathrm{\bf z}_q,\mathrm{\bf z}_g]\in\mathbb{R}^{2C}$.
The first $FC_1$ layer reduces the dimensionality $2C$ by a factor of $r$, to obtain $\frac{2C}{r}$ channels. The second layer $FC_2$ re-expands those to $C$ followed by a sigmoid activation $\sigma$. This results in the weight vector $\mathrm{s} \in \mathbb{R}^C$ being as follows
\begin{equation}
\label{eq:qseblock}
\mathrm{\bf s} = \mathrm{F}_{ex}(\mathrm{\bf z}_q,\mathrm{\bf z}_g;\mathrm{\bf W}) = \sigma(~\mathrm{\bf W}_2 ~ \delta (~\mathrm{\bf W}_1 [\mathrm{\bf z}_q, \mathrm{\bf z}_g]~)~)
\end{equation} 
whereby, $\mathrm{\bf W}_1 \in \mathbb{R}^{\frac{2C}{r} \mathrm{x} 2C}$ are the parameters of the first fully-connected layer placed for dimensionality-reduction, while the second fully-connected layer with parameters $\mathrm{\bf W}_2 \in \mathbb{R}^{C \mathrm{x} \frac{2C}{r}}$ is for dimensionality-expansion. The reduction ratio $r$ is set to 16 in all our experiments as proposed in \cite{Hu_2018_CVPR}. We refer to $\delta$ as the ReLU non-linearity that models nonlinear interactions between channels.

As shown in Fig.~\ref{fig:qseblock} (blocks \textit{``Scale''} and \textit{skip connections}), the outputs of a QSSE-ResNet block $\mathrm{\widetilde{\bf X}_Q}$ and $\mathrm{\widetilde{\bf X}_G}$ for the respective query and gallery images are:
\begin{equation}
\begin{split}
\label{eq:network}
\mathrm{\widetilde{\bf X}_Q} = \mathrm{\bf X_Q} + \mathrm{\bf s} \odot \mathrm{\bf U_Q}\\
\mathrm{\widetilde{\bf X}_G} = \mathrm{\bf X_G} + \mathrm{\bf s} \odot \mathrm{\bf U_G}
\end{split}
\end{equation} 
where $\odot$ means channel-wise multiplication, re-weighting the residual outputs $\mathrm{\bf U_Q}$ and $\mathrm{\bf U_G}$.
We connect a QSSE block to each ResNet block within the BaseNet (cf.\ Fig.~\ref{fig:network}).

Note that, differently from \cite{Hu_2018_CVPR}, our QSSE-Net concatenates globally-average-pooled features from the query and the gallery networks, and then uses the channel excitation to re-weight both of them. In this way, QSSE-Net re-calibrates channel weights to take into account \emph{intra-network} channel dependencies and \emph{inter-network} channel similarities.

\begin{figure}[t] 
\begin{center}
	\includegraphics[trim=0cm 0cm 0cm 0cm, clip=false, width=1.0\linewidth]{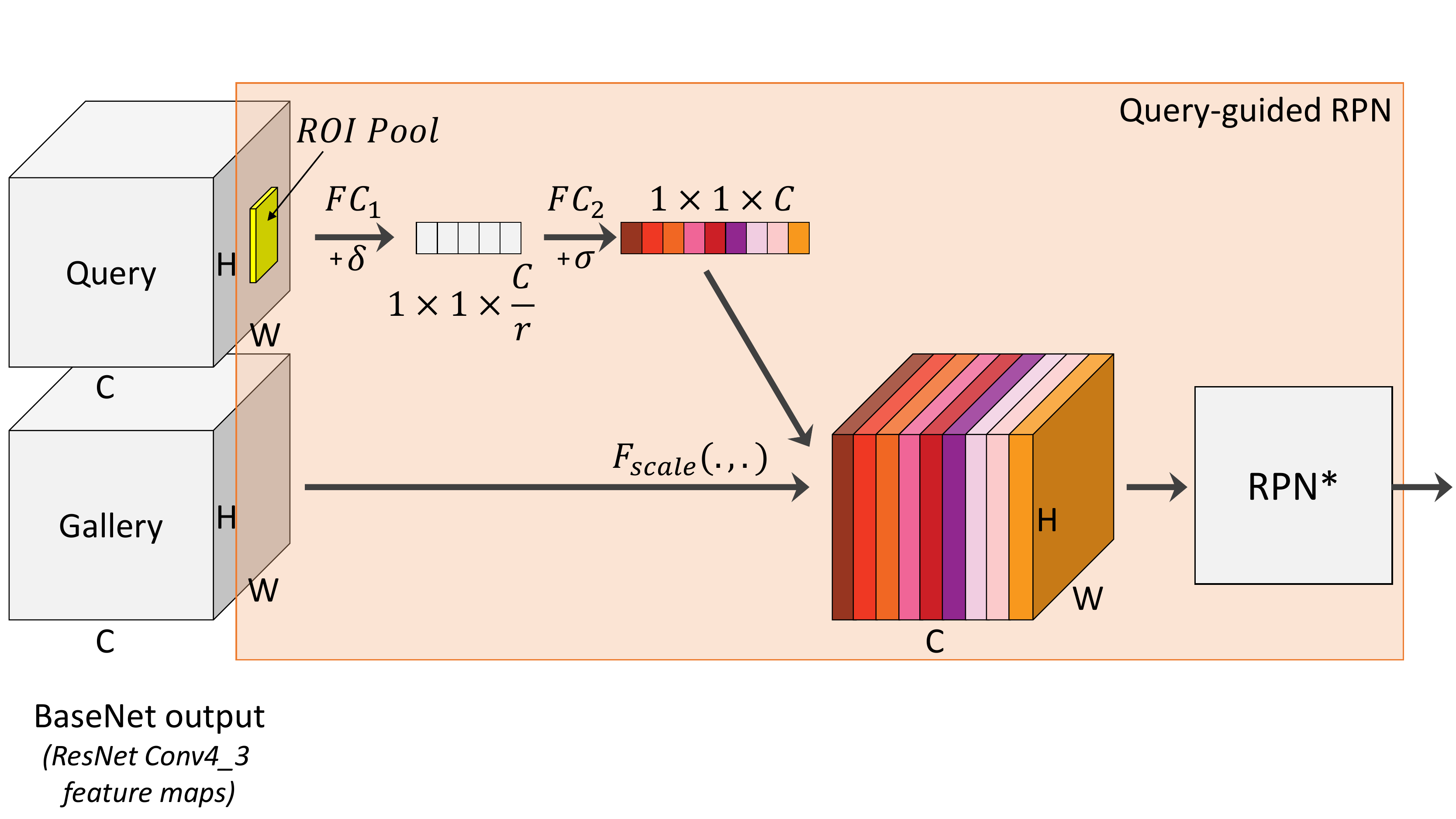}
   \vspace{-0.3cm}
	\caption{Our proposed Query-guided Region Proposal Network (QRPN). 
	Based on the query guidance, QRPN adopts a modified squeeze-and-excitation net to re-calibrate the gallery image feature responses, which are then passed to a standard RPN. (*) indicates that this RPN does not compute regression offsets.
	}\label{fig:qrpn}
   \vspace{-0.1cm}
\end{center}
\end{figure}

\subsection{Query-guided RPN (QRPN)}
\label{sec:qrpn}
Our proposed Query-guided Region Proposal Network (QRPN) re-weights the BaseNet feature responses of the gallery image by means of the cropped query features. As illustrated in Fig.~\ref{fig:qrpn}, QRPN includes a channel-wise attention mechanism (query guidance) and a standard RPN~\cite{ren2015faster}, extracting the proposal boxes from the gallery re-weighted image features. The novel query guidance is inspired by the SE block of \cite{Hu_2018_CVPR} but it features some key differences. As in \cite{Hu_2018_CVPR}, we adopt a bottleneck design with two fully-connected layers, $FC_1$ and $FC_2$, which squeeze and expand the features, so as to highlight important signal correlations. The reduction ratio $r$ is set to 16 as in \cite{Hu_2018_CVPR}. The resulting weights (excitations), upon the sigmoid activation $\sigma$, are applied to the gallery BaseNet feature maps per channel (channel-wise multiplication).

Here for the first time, we apply the SE idea to the pooled feature maps of the query crop. In more details, first we pool the query crop feature map by ROI Pool. Then we apply $FC_1$ to all channels and all pixels of the pooled map (\ie, not just to the channels). Finally, the excitations are applied to the gallery image features, not to the own query features. Our query guidance may therefore emphasize specific gallery channels, based on \emph{local} (spatially-localized) channel-wise query patterns.

Our proposed QRPN complements the standard (query-agnostic) RPN (cf.\ the parallel QRPN and RPN in Fig.~\ref{fig:network}). QRPN extracts proposal boxes featuring a \emph{query-similarity} score, while RPN pursues the standard \emph{objectness} score. 
Notably, QRPN includes an RPN with the same design as the parallel standard RPN, \eg the same anchor boxes. As illustrated in Sec.~\ref{sec:exp}, we obtain the best performance by simply summing up the scores from the QRPN and the RPN for each anchor. 
The usual non-maximum-suppression (NMS) is finally applied on the resulting score, while we adopt the regression offsets of RPN, thus query- and class-agnostic.

\subsection{Query-guided Similarity Net (QSimNet)}\label{sec:qsimnet}

The baseline OIM network~\cite{xiao2017joint} compares the re-identification features from the query and the gallery image crops by means of cosine similarity. In other words, re-id features are computed for the query and gallery image crops independently, and then used to retrieve the query individual by matching.
We maintain that the similarity score should depend on the specific query re-id features and be end-to-end trainable, so the network could emphasize and tailor the similarity metric for each query (e.g.\ balancing color, shape and other attributes for each specific person).

\begin{figure}[t] 
\begin{center}
	\includegraphics[trim=0cm 0cm 0cm 0cm, clip=false, width=1.0\linewidth]{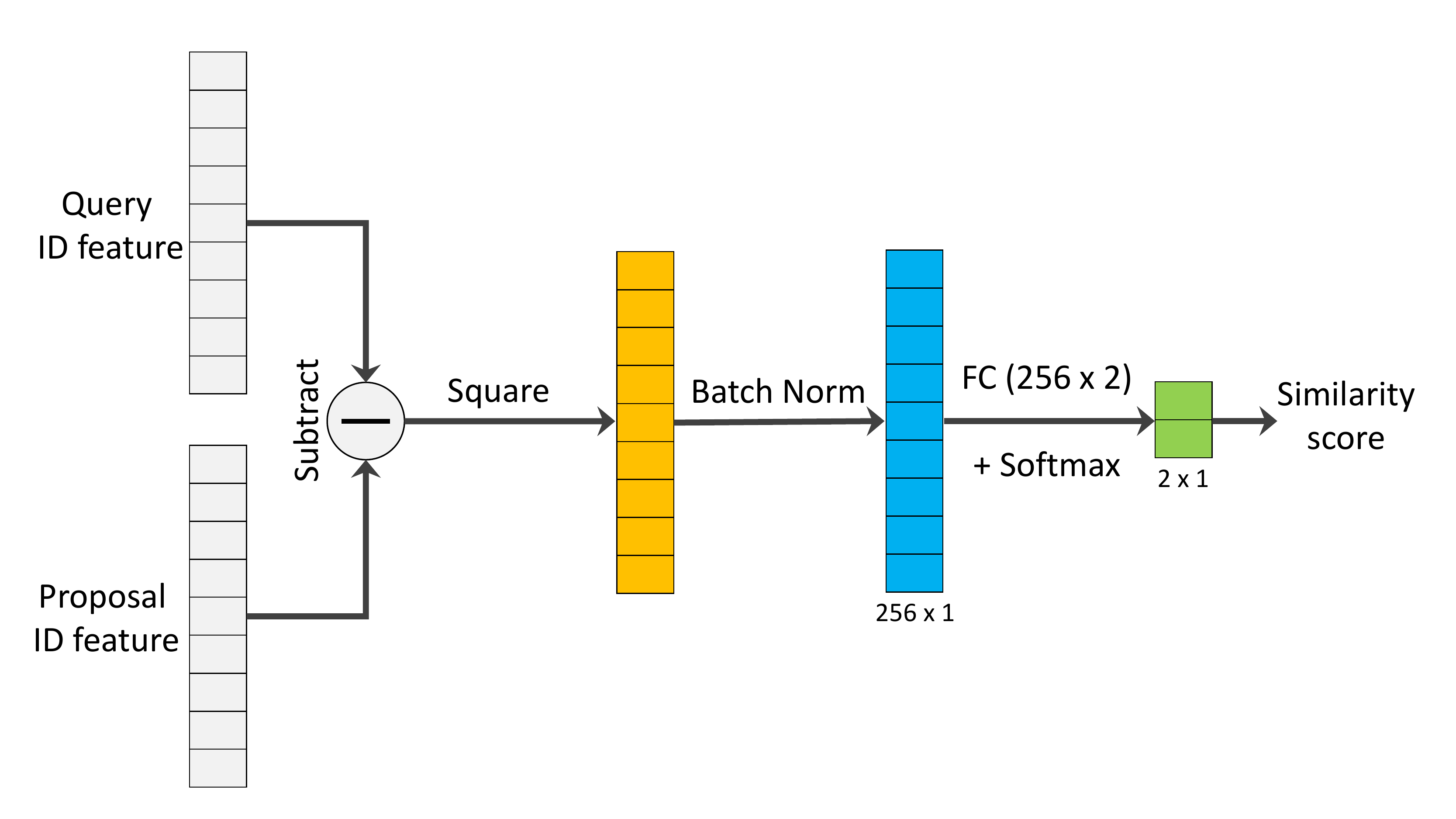}
	  \vspace{-0.5cm}
	\caption{Our proposed Query-guided Similarity Network (QSimNet).
	QSimNet introduces a simple query-guided net to estimate the similarity between the query and the gallery image. This is learned end-to-end with the rest of the network.
	}\label{fig:netsim}
    \vspace{-0.5cm}
\end{center}
\end{figure}

As illustrated in Fig.~\ref{fig:netsim}, we propose a simple query-guided similarity subnetwork (QSimNet) to compare the re-id features of the query against the gallery image crops. Upon the L2 distance (element-wise subtraction and square) of the re-id features, we apply batch normalization~\cite{Ioffe2015BatchNA} and a fully connected layer, followed by softmax.
QSimNet is learned end-to-end with the rest of the network. At inference time, we use its output scores to perform non-maximum suppression (NMS) for the final matches of the query probe in the gallery image. We do not therefore use the classification scores from the original detection network, ClsNet in Fig.~\ref{fig:network}, but ClsNet is used for training the detector branch and to remove the least-confident person detections during inference, with score~$<10^{-2}$.

\subsection{End-to-end Joint Optimization}

We jointly optimize, in an end-to-end fashion: \textbf{i.}\ the person detection network searching people in the gallery image; \textbf{ii.}\ the identification network for learning a discriminative feature embedding per ID in the training data; and \textbf{iii.}\ the novel query-guided subnetworks QSSE-Net, QRPN and QSimNet.
We pursue the joint optimization by means of loss functions for each task, represented in Fig.~\ref{fig:network} as the darker (gray or orange) loss boxes. In more details, the Faster R-CNN detector is supervised with loss functions for classification ($L_{cls}$), regression ($L_{reg}$), RPN objectness ($L_{rpn_o}$), and RPN box regression ($L_{rpn_r}$); while the identification subnetwork is supervised by the OIM loss ($L_{oim}$)~\cite{xiao2017joint}. We introduce two new loss functions, the QRPN loss ($L_{qrpn}$) and the Sim loss ($L_{sim}$), to directly supervise QRPN and QSimNet, while the QSSE-Nets gets the same implicit supervision as the BaseNets. The overall objective is given by:
\begin{equation}
\begin{split}
\label{eq:network}
L = \lambda_{1} L_{cls} &+ \lambda_{2} L_{reg} + \lambda_{3} L_{rpn_o} + \lambda_{4} L_{rpn_r}\\
&+ \lambda_{5} L_{oim} + \lambda_{6} L_{qrpn} + \lambda_{7} L_{sim}
\end{split}
\end{equation} 
whereby $\lambda_{1-7}$, responsible for the relative loss importance, are here all set to $1$.
\begin{table*}[t]
\begin{center}
\begin{tabular}{lccccc}
\hline
& \multicolumn{2}{c}{\textbf{Gallery Size 50}} & \multicolumn{2}{c}{\textbf{Gallery Size 100}}&\\
\cmidrule(l){2-3} \cmidrule(l){4-5}
Method & mAP (\%) & top-1 (\%) & mAP (\%) & top-1 (\%) & Max-recall (\%)\\
\hline
\hline
OIM \cite{xiao2017joint}  & 80.0 & - & 75.5 & 78.7 & -\\
+ \textit{QRPN} &82.1 & 82.7 & 79.6 &80.4 & 96.6 \\
+ \textit{QSimNet} & 85.1 &  85.6 & 82.6 & 83.0&  98.1\\
+ \textit{QRPN} + \textit{QSimNet}  & 86.2  & 86.7 &83.1  & 83.3 & 98.2 \\
+ \textit{QSSE} + \textit{QRPN} + \textit{QSimNet} (= QEEPS) & \bf{87.0} & \bf{ 87.1} & \bf{84.4} & \bf{84.4} & \bf{98.8}\\
\hline

\end{tabular}
\end{center}
\vspace{-0.2cm}
\caption{
Importance of each proposed model component, as evaluated on the CUHK-SYSU dataset~\cite{xiao2017joint}, for gallery sizes of 50 and 100.
OIM~\cite{xiao2017joint} results are reported from the original paper. The best performer OIM + QSSE + QRPN + QSimNet makes our proposed complete model, which we dub QEEPS. We also report the maximum detector recall, which depends on the subset of galleries containing the query.}
\vspace{-0.4cm}
\label{tab:ablation}
\end{table*}

\textbf{Siamese Design.}
Note that a query-guided person search network implies passing both the query and the gallery images through the Siamese network. Finding a person within a gallery image is still a fast operation (our Pytorch implementation runs in 300msec on an Nvidia P6000), as it only requires a forward pass through the network. But comparing a query against a gallery image set needs re-running the query at all times and, during training, it requires storing the intermediate features and the gradients for each image pair. Our current batch only contains a single query-gallery pair, but this does not affect performance (cf.\ Sec.\ref{sec:exp}).

\textbf{QRPN loss.}
Similarly to the RPN loss~\cite{ren2015faster}, we define $L_{qrpn}$ as a cross-entropy loss:
\begin{equation} \label{eq:qrpnloss}
L_{qrpn} = -\frac{1}{N}\sum_{N}\mathrm{log}(p_n^u)
\end{equation} 
whereby N is the size of the mini-batch, $p_n^u$ is the probability of the assigned true class $u$ for the $n^{th}$ anchor box in the mini-batch. Anchor boxes that overlap with the query individual are marked as positives. We set on purpose to not sample negatives from the other people present in the gallery image, to avoid setting diverging objectives for the parallel QRPN and RPN (since other people in the gallery image are positives for the standard RPN).

\textbf{Sim loss.}
We define $L_{sim}$ as the binary cross-entropy loss function which maximizes the similarity score between the query crop and the corresponding individual in the gallery image proposals. $L_{sim}$ takes similar form as Eq.~\eqref{eq:qrpnloss}.

\textbf{Positive/negative ratio.}
To alleviate for the few positives in the mini-batch (since a gallery image contains at most one query id), we augment the data via jittering and relax the IoU overlap for the anchor box positive assignment to 0.6. On the other hand, we sample fewer negatives resulting in a mini-batch of size 128 instead of 256. We keep 0.3 as the maximum IoU overlap of anchor boxes for negative assignment as in standard RPN.

\vspace{-0.1cm}
\section{Experiments}\label{sec:exp}

\vspace{-0.1cm}

\subsection{Datasets and metrics}
\label{sec:data_and_metrics}
\vspace{-0.1cm}
\textbf{CUHK-SYSU.}
The CUHK-SYSU dataset~\cite{xiao2017joint} consists of 18,184 images, labeled with 8,432 identities and 96,143 pedestrian bounding boxes (23,430 boxes are ID labeled). The images, captured in urban areas by hand-held cameras or from movie snapshots, vary largely in viewpoint, lightning, occlusion and background conditions. We adopt the standard train/test split, where 11,206 images and 5,532 identities are used for training, 2,900 queries and overall 6,978 gallery images for testing. We experiment with gallery sizes of 50 and 100 for each query. 

\textbf{PRW.}
The PRW dataset~\cite{zheng2016prw}, acquired in a university campus from six cameras, consists of 11,816 images with 43,110 bounding boxes (34,304 boxes are ID labeled)  and 932 identities. Compared to CUHK-SYSU, PRW features less images and IDs but many more bounding boxes per id (36.8, against 2.8 in CUHK-SYSU), which makes it more challenging. The training set consists of 5,134 images with 482 identities, while the test set consists of 6,112 images (gallery size) with 450 identities and provides 2057 queries.

\textbf{PRW-mini.}
The PRW test evaluation may become impractical for person search techniques which are query-based. In fact, conditioning on the query requires jointly processing each [query-gallery] pair and the exhaustive evaluation of the product space, \ie $2,057\times6,112$
\footnote{By contrast, the baseline OIM~\cite{xiao2017joint} computes query and gallery re-id features separately and requires $2,057+6,112$ network forward passes.}
.

We introduce the PRW-mini, which we publicly release\footnote{\label{footnote:web}PRW-mini and the evaluation script (for PRW and PRW-mini) are at:\\  \url{https://github.com/munjalbharti/Query-guided-End-to-End-Person-Search}}, to reduce the evaluation time while maintaining the difficulty. PRW-mini tests 30 query images against the whole gallery. To maintain difficulty, we have sampled multiple sets of 30 query images and selected the one where the baseline OIM~\cite{xiao2017joint} performs at the same accuracy as in PRW (OIM~\cite{xiao2017joint} is a de facto baseline for most recent person search techniques~\cite{Xiao2017IANTI,Liu2017NPSM}).

\textbf{Evaluation Metrics.}
We report two commonly adopted performance metrics~\cite{xiao2017joint,Xiao2017IANTI,Liu2017NPSM}: mean Average Precision (mAP) and Common Matching Characteristic (CMC top-K) for evaluation. 
mAP is derived from the detection literature and reflects the accuracy in localizing the query in all gallery images (AP is computed for each ID and averaged to compute the mAP).
CMC is specific to re-identification and reports the probability of retrieving at least one correct ID within the top-K predictions (CMC top-1 is adopted here, which we refer to as top-1).
More specifically, we evaluate on the CUHK-SYSU dataset~\cite{xiao2017joint} by using the provided scripts, and we evaluate on PRW~\cite{zheng2016prw} with the same scripts as adopted by Mask-G~\cite{Chen_2018_ECCV}, which we publicly provide\footref{footnote:web}.

\subsection{Implementation Details}\label{sec:impldetails}
We build upon OIM~\cite{xiao2017joint} for the design, setup and pre-training of the base feature network and the network head (BaseNet and ClsIdenNet in Fig.~\ref{fig:network}), as well as for the ID-Net.
As in \cite{Amin2017GeometricPF}, we adjust the anchor sizes to the objects in the dataset: we adopt scales \{2, 4, 8, 16, 32\} and aspect ratios \{1, 2, 3\}. We adopt the same anchors for the RPN and QRPN. The input images are re-scaled such that their shorter side is 600 pixels. We pad or crop the query images to the same size of the gallery one. 
We train the whole network using SGD with momentum for 2 epochs, with a base learning rate of 0.001 which is reduced by a factor of 10 after the first epoch. 
For training, we consider all query-gallery image pairs for the CUHK-SYSU dataset, but we only use three gallery images per query for the PRW dataset (since this is already large). We augment the data by flipping both the query and the gallery image.

\subsection{Ablation Study}

As ablation study, we consider the OIM~\cite{xiao2017joint} baseline and evaluate the separate benefits of the proposed QRPN, QSimNet and QSSE-Net on the CUHK-SYSU dataset.
In Table~\ref{tab:ablation}, we observe that adding QRPN to OIM provides 79.6\% mAP and 80.4\% top-1, improving mAP by 4.1pp and top-1 by 1.7pp, for a gallery size of 100. Adding QSimNet, we achieve an even higher improvement of 7.1pp mAP and 4.3pp top-1 for a gallery size of 100. Combining QRPN and QSimNet improves on both results (83.1\% mAP and 83.3\% top-1 for a gallery size of 100), demonstrating the complementary benefit of considering query guidance for the proposal generation and for the similarity score. 

We achieve the best performance (84.4\% mAP, 84.4\% top-1 for a gallery size of 100) of our proposed QEEPS network (OIM+QSSE+QRPN+QSimNet) by integrating additionally the QSSE blocks into the Siamese BaseNets. Differently from QRPN and QSimNet, QSSE-Net acts on the channels and the feature maps of the entire images, not just the local crops, which provides complementary benefits. We consider the complete QEEPS in the next experiments.

\begin{table}
\begin{center}
\begin{tabular}{lcc}
\hline
Method & mAP(\%) & top-1 (\%) \\
\hline
\hline
OIM \cite{xiao2017joint}~(Baseline)  & 75.5 & 78.7 \\
IAN \cite{Xiao2017IANTI} & 76.3 & 80.1  \\
NPSM \cite{Liu2017NPSM} & 77.9 & 81.2 \\
QEEPS (ours) & \textbf{84.4} & \textbf{84.4} \\
\hdashline[3pt/5pt]
Mask-G \cite{Chen_2018_ECCV} & 83.0 & 83.7 \\
OIM\ddag~(Baseline) & 83.3 & 84.2 \\
QEEPS (ours)  & \bf{88.9} & \bf{89.1} \\
\hline
\end{tabular}
\end{center}
\vspace{-0.2cm}
\caption{Comparison with the state-of-the-art on the CUHK-SYSU dataset for the gallery size 100. Methods above the dashed line employ the standard Faster R-CNN image re-sizing to 600 pixels (shorter image side), while those below use larger images with shorter sides of 900 pixels. Note the strength of our baseline OIM\ddag, which is already above the state-of-the-art performance.}
\vspace{-0.5cm}
\label{tab:sota_cuhk}
\end{table}

In the last column of Table~\ref{tab:ablation}, we show a similar improvement trend for the detector recall (for a fixed number of region proposals of 300). 
Our full model QEEPS achieves a nearly-perfect recall of 98.8\%, indicating the importance of query guidance also for learning higher-quality proposals.




\subsection{Comparison to the State-of-the-art} \label{sec:sota}

\paragraph{CUHK-SYSU.}
In Table~\ref{tab:sota_cuhk}, we compare QEEPS to state-of-the-art approaches in person search \cite{Chen_2018_ECCV,Liu2017NPSM,Xiao2017IANTI} and to  OIM~\cite{xiao2017joint}. It should be noted that OIM~\cite{xiao2017joint}, IAN~\cite{Xiao2017IANTI} and NPSM~\cite{Liu2017NPSM} build on Faster R-CNN and therefore presumably
re-scale the images such that shorter side is 600 pixels. All three methods (shown above the \textit{dashed} lines) argue for a joint detection and re-identification network. For this image resolution, our approach QEEPS achieves 84.4\% mAP and 84.4\% top-1, surpassing the state-of-the-art NPSM \cite{Liu2017NPSM} by 6.5pp mAP and 3.2pp top-1 respectively.   

Below the \textit{dashed} line, Mask-G \cite{Chen_2018_ECCV} considers a similar Faster R-CNN but with larger images (shoter side 900 pixels). In order to have a fair comparison with Mask-G~\cite{Chen_2018_ECCV}, we consider the baseline OIM\ddag\,, same as OIM but with the input images re-scaled to a shorter side of 900 pixels, as long as the larger side be less than 1500 pixels. Mask-G argues that its better performance (83.0\% mAP, 83.7\% top-1 for a gallery of 100) is due to considering detection and re-identification independently. However, when run on the larger images, the strong baseline OIM\ddag\ surpasses all prior art with a performance of 83.3\% mAP and 84.2\% top-1. We argue that this reasserts the validity of considering detection and re-identification jointly. On the same setup, our QEEPS achieves 88.9\% mAP and 89.1\% top-1, improving on best published results (Mask-G) by 5.9pp mAP and 5.4pp top-1. We attribute the further leap in performance to the proposed query guidance, both as a global and local cue.

\vspace{-0.5cm}
\begin{table}
\begin{center}
\begin{tabular}{lcc}
\hline
Method & mAP(\%) & top-1 (\%) \\
\hline
\hline
OIM \cite{xiao2017joint}  & 21.3 & 49.9 \\
IAN \cite{Xiao2017IANTI} & 23.0 & 61.9  \\
NPSM \cite{Liu2017NPSM} & 24.2 & 53.1 \\
Mask-G \cite{Chen_2018_ECCV} & 32.6 & 72.1 \\
OIM\ddag\ (Baseline) &  36.9 & 75.7\\
QEEPS (ours)& \bf{37.1} & \bf{76.7} \\
\hdashline[3pt/5pt]
Mask-G \cite{Chen_2018_ECCV} & 33.1 & 70.0 \\
OIM\ddag\ (Baseline) & 38.3 & 70.0 \\
QEEPS (ours)& \bf{39.1} & \bf{80.0} \\
\hline
\end{tabular}
\end{center}
\vspace{-0.2cm}
\caption{Comparison with the state-of-the-art on the PRW dataset~\cite{zheng2016prw}, above the dashed line, and on the proposed subset PRW-mini (cf.\ Sec.~\ref{sec:exp}), below it.}
\label{tab:sota_prw}
\vspace{-0.5cm}
\end{table}

\tabcolsep 1.0pt
\renewcommand{\arraystretch}{0.5}

\newlength{\cuhkfigheight}
\setlength{\cuhkfigheight}{1.5cm}

\begin{figure*}[t]
\begin{center}
\begin{tabular}{ccc|ccc}
\includegraphics[trim=0cm 0cm 0cm 0cm, clip=true, height=\cuhkfigheight]{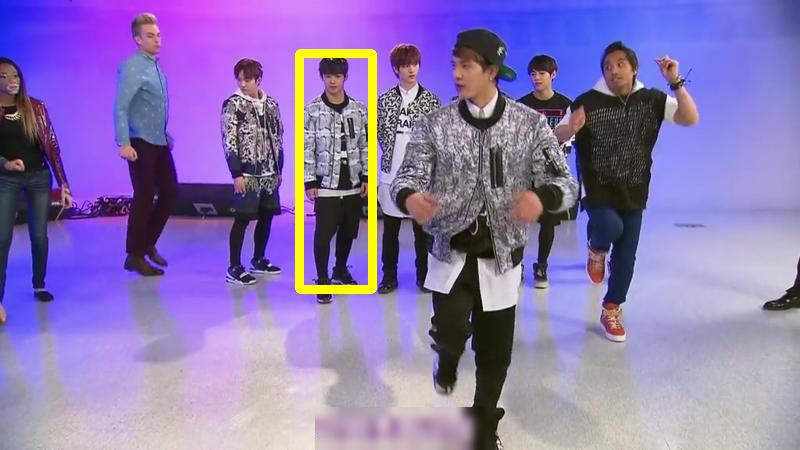}&
\includegraphics[trim=0cm 0cm 0cm 0cm, clip=true, height=\cuhkfigheight]{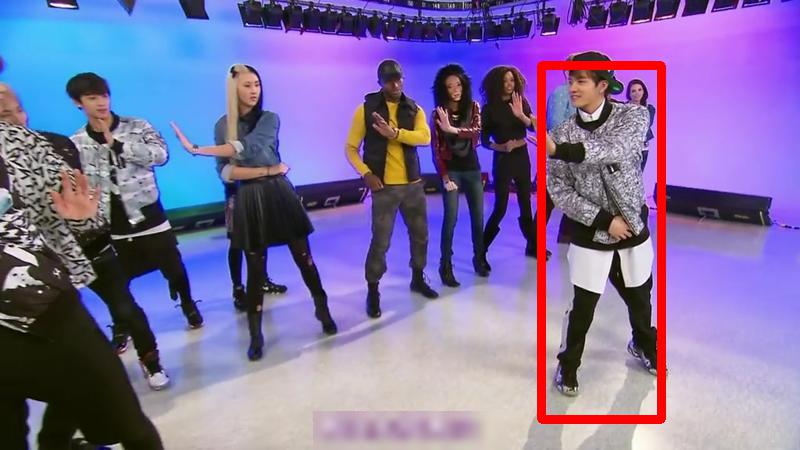}&
\includegraphics[trim=0cm 0cm 0cm 0cm, clip=true, height=\cuhkfigheight]{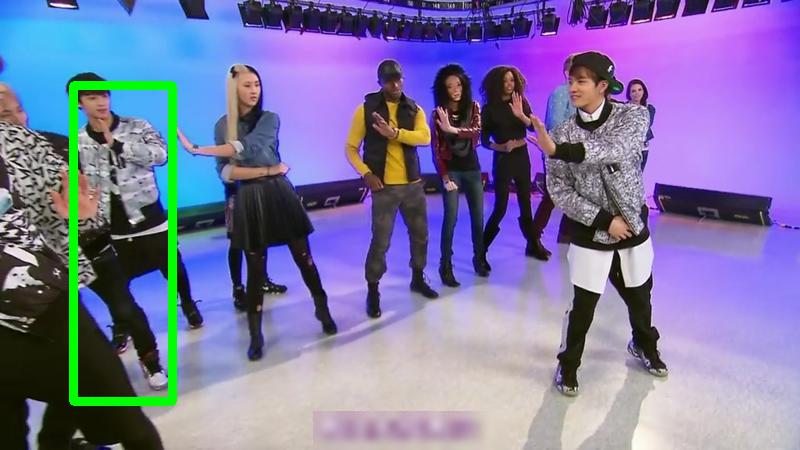}&
\includegraphics[trim=0cm 0cm 0cm 0cm, clip=true, height=\cuhkfigheight]{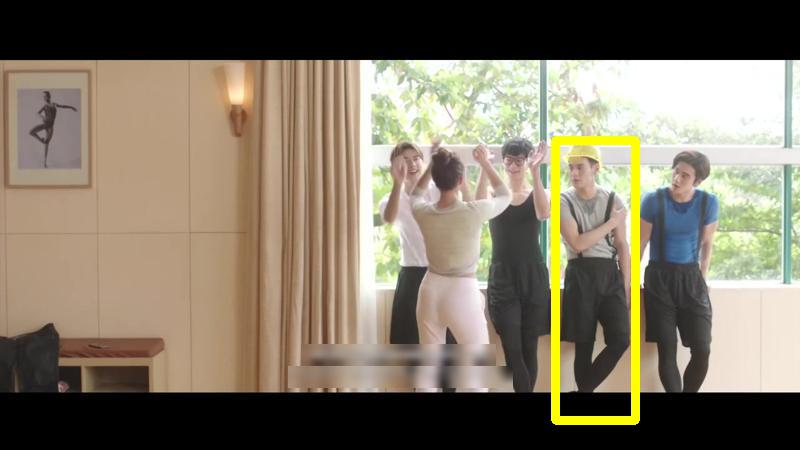}&
\includegraphics[trim=0cm 0cm 0cm 0cm, clip=true, height=\cuhkfigheight]{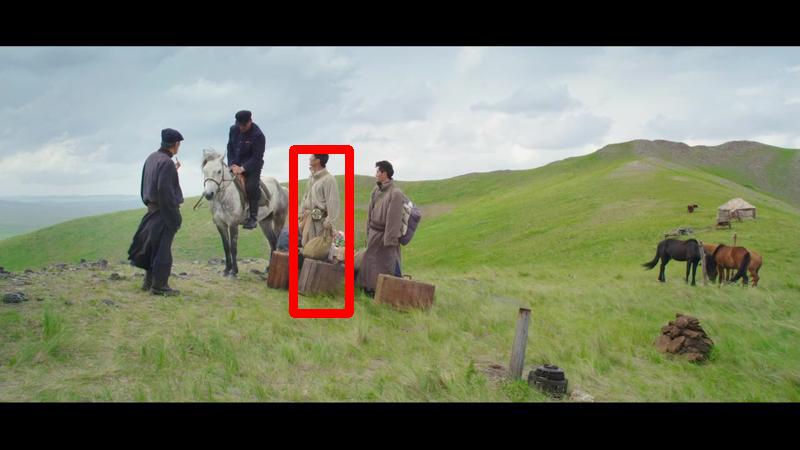}&
\includegraphics[trim=0cm 0cm 0cm 0cm, clip=true, height=\cuhkfigheight]{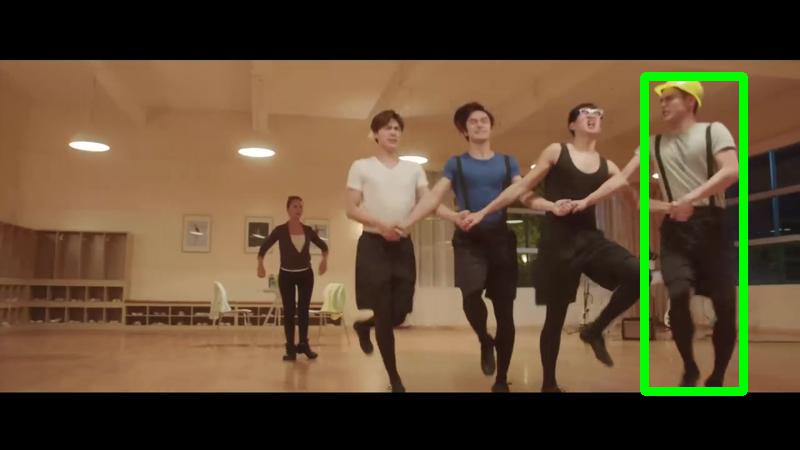}\\

\includegraphics[trim=0cm 0cm 0cm 0cm, clip=true, height=\cuhkfigheight]{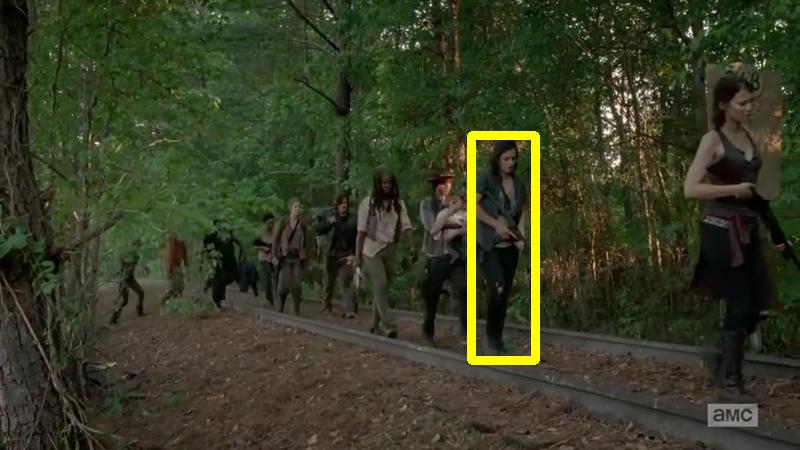}&
\includegraphics[trim=0cm 0cm 0cm 0cm, clip=true, height=\cuhkfigheight]{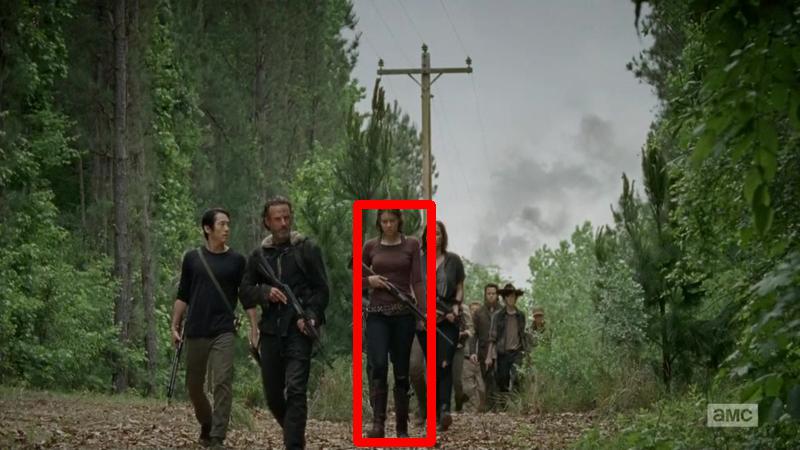}&
\includegraphics[trim=0cm 0cm 0cm 0cm, clip=true, height=\cuhkfigheight]{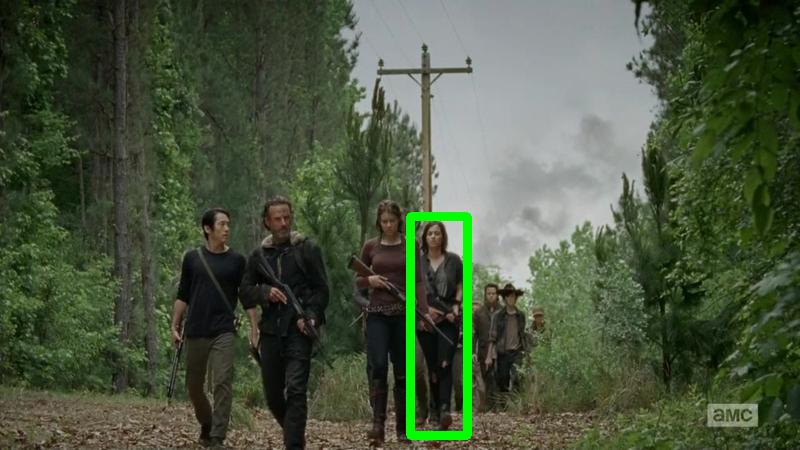}&
\includegraphics[trim=0cm 0cm 0cm 0cm, clip=true, height=\cuhkfigheight]{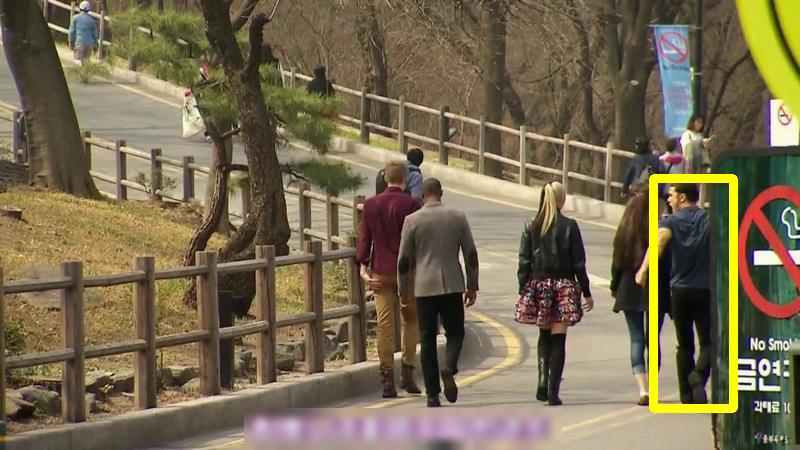}&
\includegraphics[trim=0cm 0cm 0cm 0cm, clip=true, height=\cuhkfigheight]{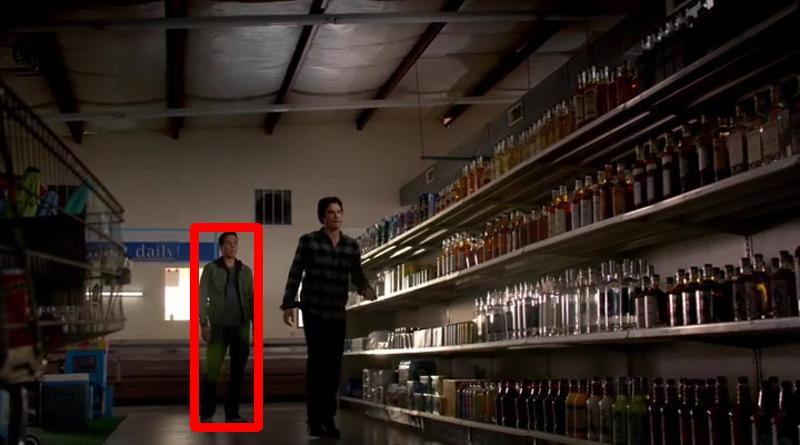}&
\includegraphics[trim=0cm 0cm 0cm 0cm, clip=true, height=\cuhkfigheight]{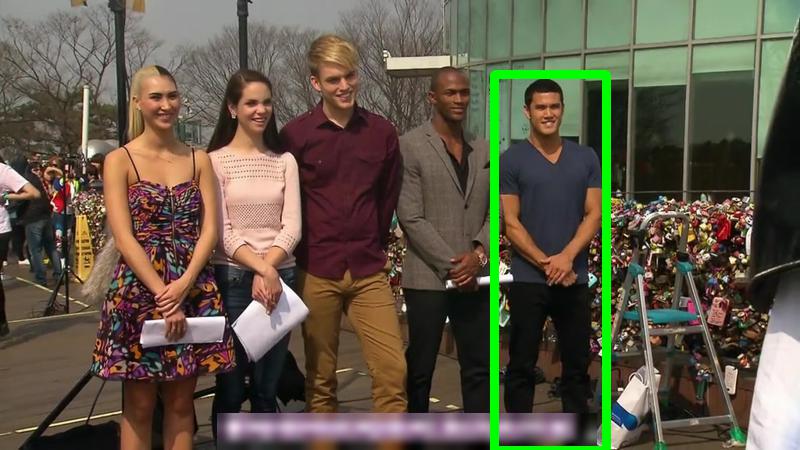}\\

\includegraphics[trim=0cm 2cm 2cm 2cm, clip=true, height=\cuhkfigheight]{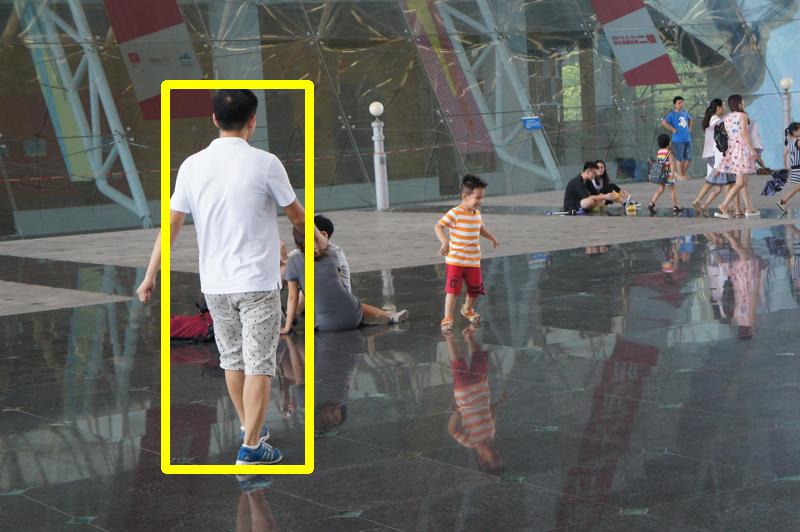}&
\includegraphics[trim=0cm 0cm 0cm 5cm, clip=true, height=\cuhkfigheight]{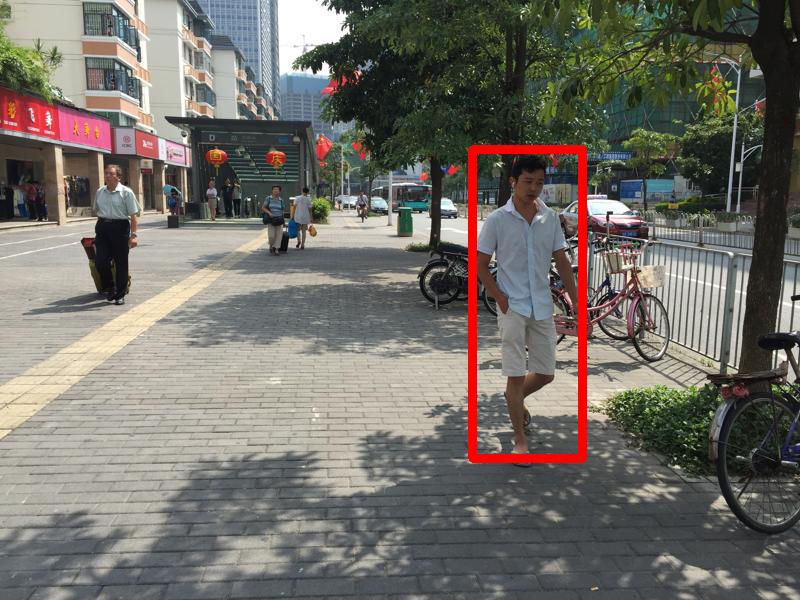}&
\includegraphics[trim=0cm 3cm 0cm 0cm, clip=true, height=\cuhkfigheight]{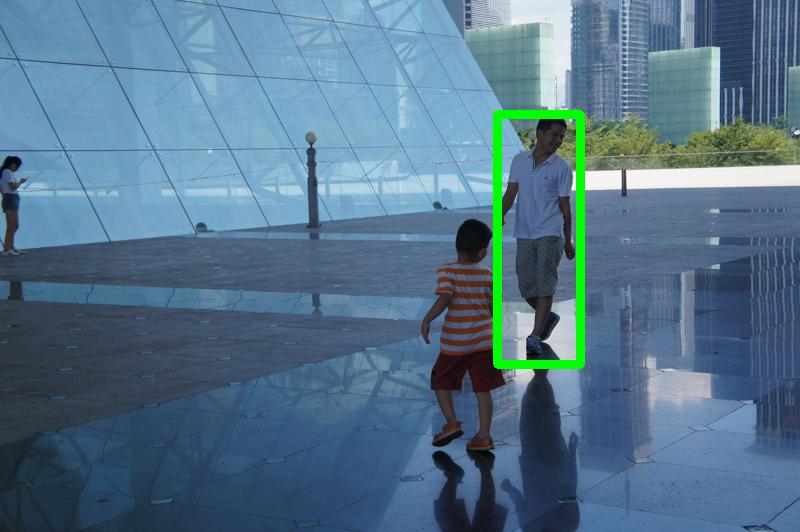}&

\includegraphics[trim=0cm 13cm 0cm 3.3cm, clip=true, height=\cuhkfigheight]{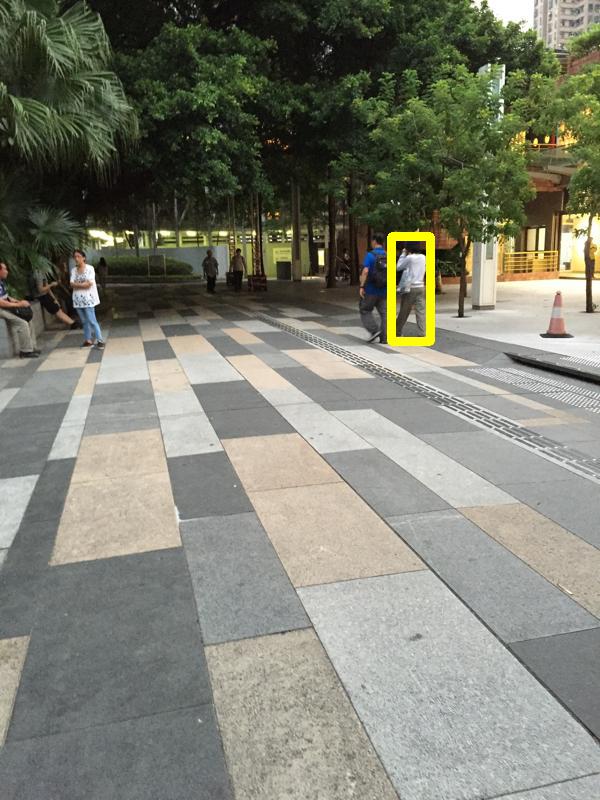}&
\includegraphics[trim=0cm 0.3cm 0cm 5cm, clip=true, height=\cuhkfigheight]{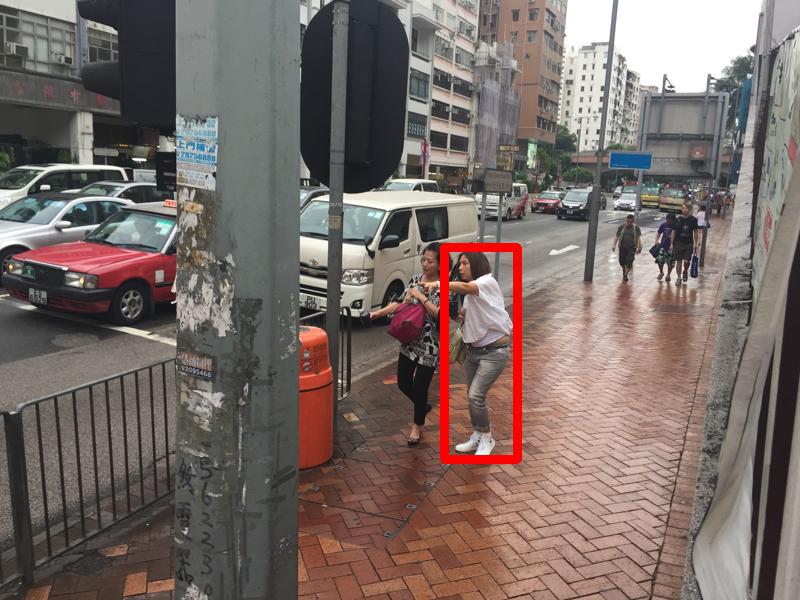}&
\includegraphics[trim=0cm 13cm 0cm 3.3cm, clip=true, height=\cuhkfigheight]{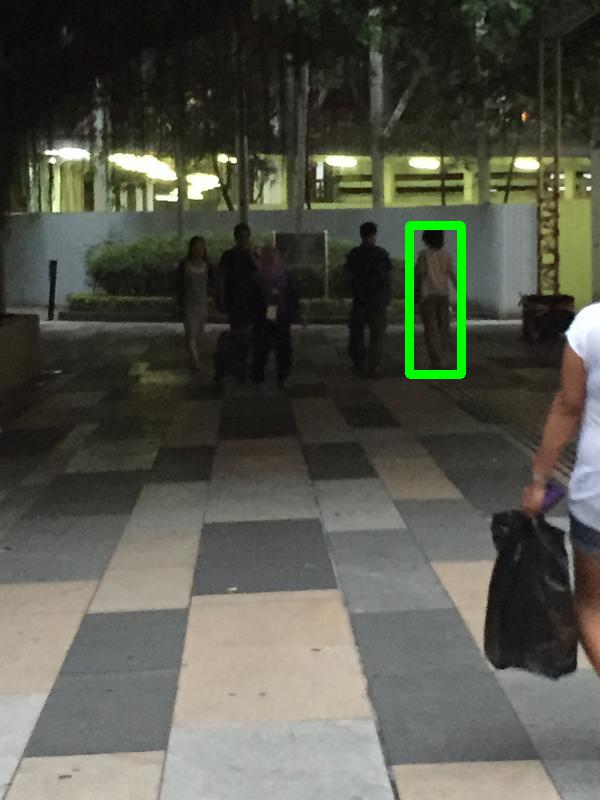}\\

\cmidrule(r){1-3} \cmidrule(l){4-6}
(a) Query & (b) OIM\ddag\ Top-1 & (c) QEEPS Top-1 & (a) Query & (b) OIM\ddag\ Top-1 & (c) QEEPS Top-1

\end{tabular}
\vspace{0.1cm}
\caption{
Qualitative \emph{Top-1} person search results for a number of challenging query examples.
For each example, we show (a) the query images with the bounding box of the query-person, in yellow, (b) their corresponding output matches given by the baseline OIM\ddag\ ,
and (c) results of our proposed approach QEEPS. 
Notice that, our approach is able to fetch difficult gallery images as its first estimate.
Red bounding boxes are failures, while green represent correct matches.
}
\vspace{-0.5cm}
\label{fig:results}
\end{center}
\end{figure*}


\tabcolsep 1.0pt
\renewcommand{\arraystretch}{0.5}

\newlength{\cuhkfigheightfail}
\setlength{\cuhkfigheightfail}{1.6cm}

\begin{figure*}[t]
\begin{center}
\begin{tabular}{cc|cc|cc}
\includegraphics[trim=0cm 0cm 0cm 0cm, clip=true, height=\cuhkfigheightfail]{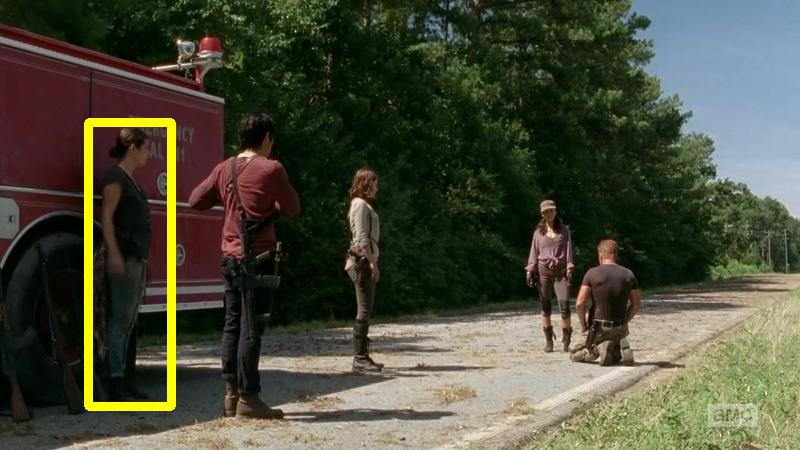}&
\includegraphics[trim=0cm 0cm 0cm 0cm, clip=true, height=\cuhkfigheightfail]{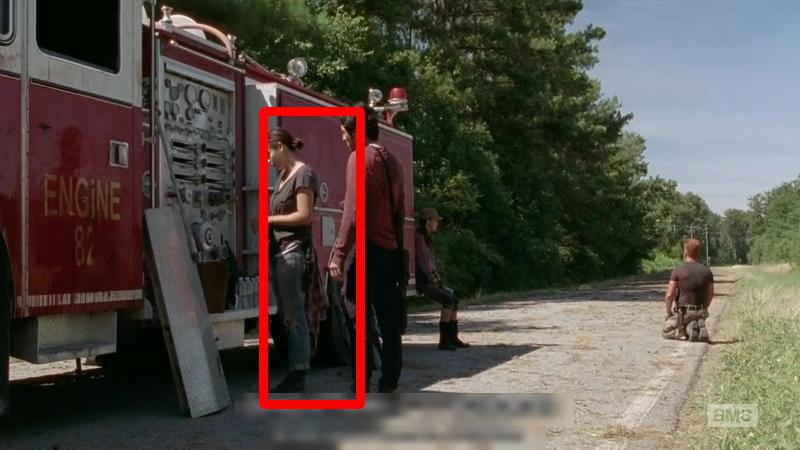}&
\includegraphics[trim=0cm 0cm 4cm 0cm, clip=true, height=\cuhkfigheightfail]{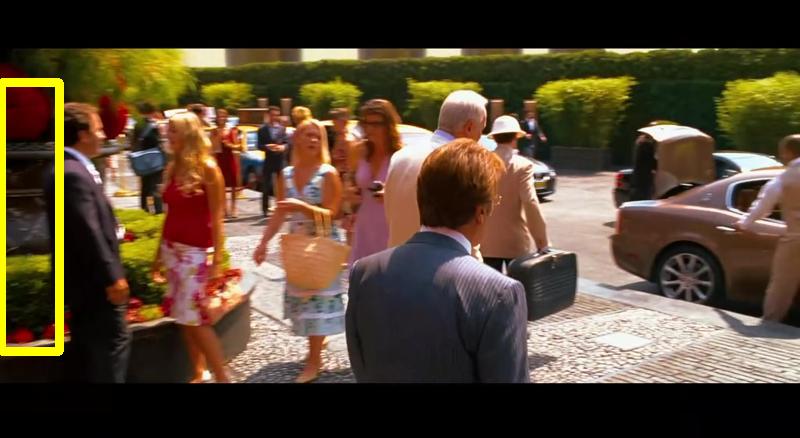}&
\includegraphics[trim=0cm 6cm 0cm 6cm, clip=true, height=\cuhkfigheightfail]{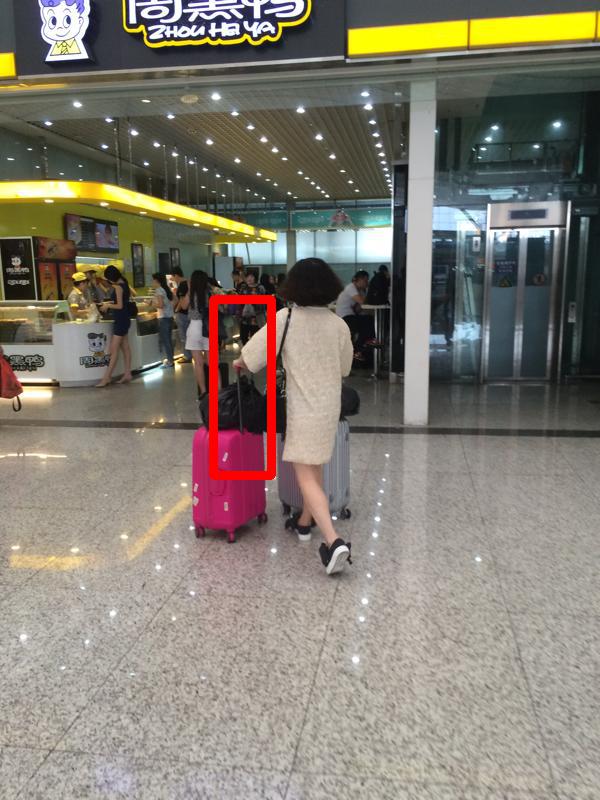}&
\includegraphics[trim=0cm 0cm 0cm 0cm, clip=true, height=\cuhkfigheightfail]{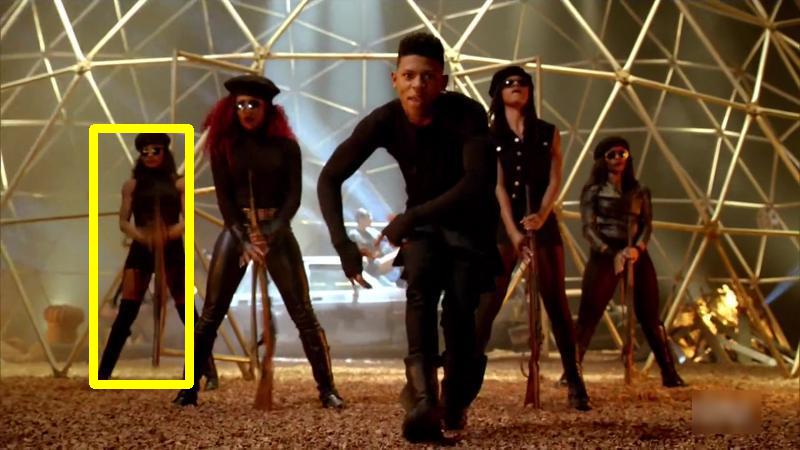}&
\includegraphics[trim=0cm 0cm 0cm 0cm, clip=true, height=\cuhkfigheightfail]{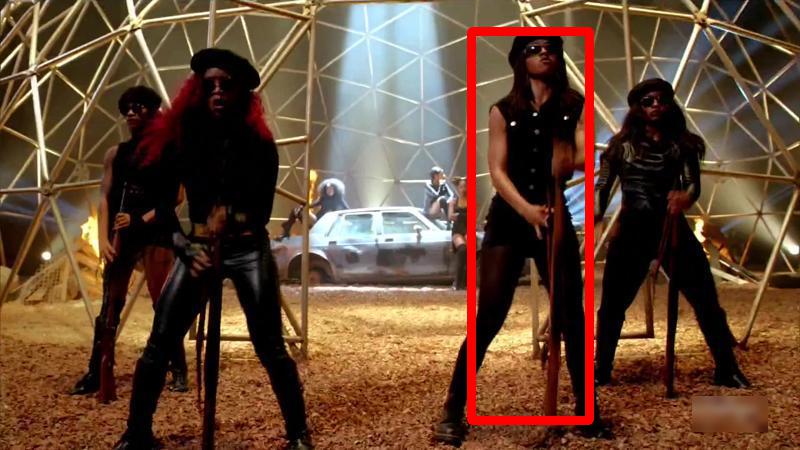}\\
Query & QEEPS Top-1 & Query & QEEPS Top-1 & Query & QEEPS Top-1\\
\multicolumn{2}{c}{(a)} & \multicolumn{2}{c}{(b)} & \multicolumn{2}{c}{(c)}
\end{tabular}
\caption{
Observed trends in failure cases, such as,
(a) localization errors of the person detector,
(b) incorrect ground-truth annotations of bounding box and the person ID,
and (c) extremely challenging examples due to similar appearance and/or low visibility.
}
\vspace{-0.6cm}
\label{fig:results_fail}
\end{center}
\end{figure*}


\paragraph{PRW/PRW-mini.}
In Table~\ref{tab:sota_prw}, we compare state-of-the-art techniques~\cite{Chen_2018_ECCV,Liu2017NPSM,Xiao2017IANTI} to OIM~\cite{xiao2017joint}, to the baseline OIM\ddag\ and to the proposed QEEPS. Results above the dashed line refer to the full PRW dataset~\cite{zheng2016prw}. Note how Mask-G~\cite{Chen_2018_ECCV} (32.6\% mAP, 72.1\% top-1) and OIM\ddag\ (36.9\% mAP, 75.7\% top-1) neatly surpass all other approaches. While one cannot draw a clear conclusion on the employed technique, it seems clear that processing larger input images yields a strong benefit, since these two methods are the only to re-scale them to short sides of 900 pixels, instead of 600. Finally, QEEPS outperforms the Mask-G by
4.5pp in mAP and 4.6pp in top-1, setting the novel state-of-the-art performance of 37.1\% mAP and 76.7\% top-1.

Below the dashed line, we report results on the PRW-mini subset, introduced in Sec.~\ref{sec:data_and_metrics}. Note how the ranking of OIM\ddag\,, Mask-G~\cite{Chen_2018_ECCV} and QEEPS is preserved in the PRW-mini as compared to PRW, and that all algorithms report similar mAP and top-1 performances (\eg for Mask-G the gaps are only 0.5pp and 2.1pp respectively). PRW-mini maintains therefore the same difficulty as PRW, while reducing the evaluation time for query-based techniques by 2 orders of magnitude.

\vspace{-0.5cm}

\paragraph{Runtime Comparison.}
In Table~\ref{tab:complexity}, we report the time taken by Mask-G  to
process a gallery image (given a pre-processed query) compared to ours (processing both query and gallery images at
all times). Since both methods use different GPUs, we report the TFLOPs too. Upon normalization with TFLOPs, ours is ∼3.14 times faster and requires ∼4.27 times less memory.


\begin{table}[h]
\begin{center}
\begin{tabular}{l c c c}
\hline
Method  \hspace{0.5pc} & \# params (\textit{M}) \hspace{0.5pc} & Time (\textit{sec}) \hspace{0.5pc} & GPU (TFLOPs)  \\
\hline
\hline
Mask-G~\cite{Chen_2018_ECCV}  & 209 & 1.3 & K80 (8.7) \\
QEEPS & 49 & 0.3 & P6000 (12.0) \\
 \hline
\end{tabular}
\end{center}
\vspace{-0.2cm}
\caption{Runtime comparison of QEEPS with Mask-G [4] for image size $900 \times 1500$.}
\vspace{-0.3cm}
\label{tab:complexity}
\end{table}

\subsection{Qualitative Results}

As illustrated in Fig.~\ref{fig:results}, our approach performs person search successfully in a number of challenging scenarios, where the baseline method fails.
For instance, in the top-left example, QEEPS retrieves the correct guy dancing in the line, while OIM\ddag\ selects a different individual, but similarly dressed. Also quite convincingly, in the middle-right example, the query person is provided from the back (at low resolution) and found in a frontal-view gallery image. In the last row, we show two interesting examples depicting the importance of the global context. Notice how QEEPS compensates for global illumination changes (\eg the blueish image in the bottom-left example) and retrieves the correct person.

We show in Fig.~\ref{fig:results_fail} the three most common failure cases. In column (a), the same person is retrieved successfully but wrongly localized (IoU $< 0.5$). In column (b), we illustrate an annotation mistake. Finally in column (c) the failure is most likely due the extreme difficulty of some examples on the CUHK-SYSU and PRW datasets, since several people look alike (illustrated) and some others have low-visibility issues. These may be challenging to a human, too.


\section{Conclusions}

We have proposed a novel QEEPS network, which jointly addresses detection and re-identification in an end-to-end fashion.
Our results demonstrate that the joint consideration of detection and re-identification is a valid approach to person search, as it intuitively allows each separate module to account and to change with each other, during the joint training.
Furthermore, the large and consistent improvement in performance provided by our proposed query guidance highlights the importance of this aspect. When searching for a person in a gallery image, we should consider the query for its global context (e.g.\ the overall illumination may shift the importance of color as a cue) and for its local cues (e.g.\ specific patterns which ease the creation of tailored proposals and better similarity scores).

\vspace{0.4cm}

\noindent\textbf{Acknowledgements}
This research was partially funded by the BMWi -- German Federal Ministry for Economic
Affairs and Energy (MEC-View Project). We are grateful to the Mask-G team for their great support on the PRW evaluation.


{\small
\bibliographystyle{ieee}
\bibliography{egbib}
}

\end{document}